%% file: arxiv.tex
\documentclass{article}
\usepackage{arxiv}

\usepackage[backref=page]{hyperref}
\usepackage[square,sort,comma,numbers]{natbib}
\usepackage{siunitx,etoolbox}  
\usepackage[utf8]{inputenc} 
\usepackage[T1]{fontenc}    
\usepackage{url}            
\usepackage{booktabs}       
\usepackage{amsfonts}       
\usepackage{nicefrac}       
\usepackage{microtype}      
\usepackage{lipsum}		
\usepackage{graphicx}
\usepackage{doi}
\usepackage{tikz}
\usepackage{comment}
\usepackage{amsmath,amssymb}
\usepackage{color}
\usepackage{flafter} 
\usepackage{threeparttable}

\begin{document}
\pagestyle{headings}
\title{Task Specific Attention is one more thing you need for object detection}

\author{Lee, Sang yon \\
%
 NAVER Vision \\
 syon.lee@navercorp.com}
\maketitle
\input{document.tex}
\end{document}

%% file: document.tex
\begin{abstract}
Various models have been proposed to perform object detection. However, most require many hand-designed components such as anchors and non-maximum-suppression(NMS) to demonstrate good performance. 
To mitigate these issues, Transformer-based DETR \cite{DBLP:journals/corr/abs-2005-12872_DETR} and its variant, Deformable DETR \cite{Zhu2020DDETR}, were suggested. These have solved much of the complex issue in designing a head for object detection models; however, doubts about performance still exist when considering Transformer-based models as state-of-the-art methods in object detection for other models \cite{Zhang2020ATSS, chen2020reppointsv2} depending on anchors and NMS revealed better results. Furthermore, it has been unclear whether it would be possible to build an end-to-end pipeline in combination only with attention modules, because the DETR-adapted Transformer method used a convolutional neural network (CNN) for the backbone body. In this study, we propose that combining several attention modules with our new Task Specific Split Transformer (TSST) is a powerful method to produce the state-of-the art performance on COCO results without traditionally hand-designed components. By splitting the general-purpose attention module into two separated goal-specific attention modules, the proposed method allows for the design of simpler object detection models. Extensive experiments on the COCO benchmark
demonstrate the effectiveness of our approach. Code is available at https://github.com/navervision/tsst
\end{abstract}

\section{Introduction}
\begin{center}
\input{figure1}
\end{center}
Deep learning models have been transforming from inductive-biased models (e.g. CNNs \cite{DBLP:journals/corr/HeZRS15, DBLP:journals/corr/XieGDTH16}), to more flexible models (e.g. Transformers \cite{Vaswani2017TR}). Computer vision tasks including classification and object detection are also subject to this trend.  Vision transformer backbones such as Vit\cite{DBLP:journals/corr/abs-2010-11929_ViT}, DeiT\cite{DBLP:journals/corr/abs-2012-12877_DeiT} and Swin Transformer \cite{Liu2021Swin} transform traditional CNN-based backbones into more flexible and higher-performance Transformer-based backbones. DETR \cite{DBLP:journals/corr/abs-2005-12872_DETR} and Deformable DETR \cite{Zhu2020DDETR} encompass transforming the hand-designed detection heads to the Transformer-based detection heads.

However, we have not found a study that unifies Transformer building blocks in most of the parts with the aim of developing a superior end-to-end Transformer model: In aforementioned cases, (1) DETR and Deformable DETR were attached only to the traditional CNN backbones and they were not tested with Transformer-based backbones. (2) Transformer backbones have been proved their effectiveness when fused with the other detection heads but Transformer.\cite{Liu2021Swin, Dai_2021_CVPR_DyHead} Superiority of dual Transformers when used in backbones and detection heads  simultaneously is still a mission to be proved. 

In the object detection problem, it comprises two separate cost measures, which are classification and localization. Even though Transformers have showed their effectiveness for multiple subjects for each case, we found out separated Transformer head blocks for each specific task is more effective than original shared Transformer head blocks. Based on this finding, we suggest Task Specific Split Transformer (TSST) which supports separate Transformer block for each task.

In this study, as illustrated in Figure \ref{fig:fig1}, we propose a novel combination of the Transformer-based backbone and Transformer-based head with additional Transformer modules that specialize at classification and localization. We found that simple fusing between Transformer backbone and Transformer detection head is a good baseline to begin with. Further, we aim to prove that adding our specialized TSST heads is the superior compromise between a simple structure and performance. 

We evaluate TSST on the popular object detection benchmark COCO \cite{DBLP:journals/corr/LinMBHPRDZ14_COCO} to demonstrate its effectiveness. With adding only 7\% more parameters to Deformable DETR, TSST shows significantly better performance than Deformable DETR and recently released other models. \cite{Zhang2020ATSS, chen2020reppointsv2, Dai_2021_CVPR_DyHead} Code is available at https://github.com/navervision/tsst.

\section{Related Work}
\label{sec:headings}

\subsection{Backbones for Object Detection}
\subsubsection{CNN models}
Object detection models, including Transformer-headed detectors such as DETR and Deformable DETR, have relied on CNN-based ResNet backbone variants, such as ResNet-50, ResNet-101 \cite{DBLP:journals/corr/HeZRS15}, ResNext-101 \cite{DBLP:journals/corr/XieGDTH16}, ResNext-101-DCN \cite{DBLP:journals/corr/DaiQXLZHW17}, which have been the de-facto performance comparison scheme between detection methods before attention-based backbones has newly introduced. For fair and consistent comparison between older and newer detection models, CNN-based ResNet backbones are still used to validate object detection benchmarks.  

\subsubsection{Attention models}
\paragraph{Transformer.} Contrary to convolutional neural networks, Transformers \cite{Vaswani2017TR} use attention to model long-range dependencies in the data. Therefore, with a proper modeling and training strategy, models for vision tasks that have replaced CNNs with Transformers are expected to achieve better performance.

Because the pixel space of images is far larger than the sequence space of natural languages, it seemed next to impossible to construct the self- or cross-attention between all pixels in high resolution images due to computational complexity with quadratic cost in the number of pixels. This was the essential problem to design good Transformer backbone models.

\paragraph{Vision Transformers(ViT)} ViT \cite{DBLP:journals/corr/abs-2010-11929_ViT} is one of the initial studies to attempt to implement a language-task-based Transformer into the visual-task-targeted backbone structure. First, ViT followed the encoder structure of the original Transformer \cite{Vaswani2017TR}; however, it uses visual embedding instead of word embedding. Second, because there is no sequence in the image inputs, it splits an image’s raw features into several patches and used them as patch embeddings. In addition, it uses 2-D positional embedding because image features exist in 2-dimensional space. Even though ViT is a good initial try to replace CNNs with Transformers, the performace of ViT did not stand out, compared to CNNs. After ViT was introduced, to surpass the performance of CNNs and ViT, other Transformer \cite{DBLP:journals/corr/abs-2012-12877_DeiT, Liu2021Swin}-based backbones have been developed by using the structure of ViT as reference.

\paragraph{Swin Transformer.} After some efforts, meticulously engineered Transformers have surpassed CNN competitors in vision task benchmarks. Swin Transformers \cite{Liu2021Swin}, which are one of the most successful Transformer-based backbones to date, reported better results in the COCO object detection benchmark as well as in the classification benchmark. The Swin Transformer \cite{Liu2021Swin} constructs hierarchical feature maps and has computational complexity linear to image size. It constructs a hierarchical representation
by starting from small patches and gradually merging neighboring patches in deeper
Transformer layers. The Swin Transformer is constructed using windows that are arranged such that the image is evenly partitioned without overlapping. The number of patches in each window is fixed, and thus, the complexity becomes linear to the image size. These merits make the Swin Transformer suitable as a general-purpose backbone for various vision tasks. The non-overlapping windows in the Swin Transformer shift in each layer, creating cross-window connections.
\subsection{Object Detection}

\subsection{Conventional Models for Object Detection}
\paragraph{2-stage models}
Deep-learning-based object detection methods are mainly classified into two categories: 1-stage models \cite{ DBLP:journals/corr/abs-1708-02002, DBLP:journals/corr/RedmonDGF15} and 2-stage models \cite{DBLP:journals/corr/RenHG015_FasterRCNN, DBLP:journals/corr/abs-1712-00726_CascadeRCNN}. In 2-stage models, the first stage is used to extract object regions (region of interests, ROIs \cite{DBLP:journals/corr/RenHG015_FasterRCNN}), and the second stage is used to classify and further refine the localization of the object. In 1-stage models, there is no dedicated process to extract region of interests, but models directly refine the class and location of objects. 

The well-known Faster R-CNN \cite{DBLP:journals/corr/RenHG015_FasterRCNN} and Faster-R-CNN-based Cascade R-CNN \cite{DBLP:journals/corr/abs-1712-00726_CascadeRCNN} 2-stage object detection models have been the best-performing object detection models.
These may remain the best models if reinforced with more powerful backbones and other more recently developed techniques. \cite{Liu2021Swin} Their effectiveness has been proved in the public competition OpenImages \cite{OpenImages}. Top-ranked participants \cite{DBLP:journals/corr/abs-2003-07540, zhou2021simple} have, at their own discretion, selected Cascade R-CNN or its close variants as their primary choice. 
However, even if it is the best performing detection model, the development of more efficient models is being considered due to the relatively high cost of computation for inference. 

\paragraph{1-stage models}
1-stage object detection models are supposedly simple and fast; however, their performance in terms of well-known mean average precision (MAP) scores is not good enough to compare with 2-stage models. 
However, RetinaNet \cite{DBLP:journals/corr/abs-1708-02002} and Yolo \cite{DBLP:journals/corr/RedmonDGF15} variants have gradually advanced, rapidly narrowing the performance gap between 2-stage and 1-stage models.
Nevertheless, dissatisfaction related to 1-stage model structure still exists because they still heavily depend on hand-designed components, such as anchors and a non-maximum-suppression threshold, which are expected to be removed in the future to allow for simpler and more intuitive models.

\paragraph{Detection Head Add-ons}
In addition to the 2-stage or 1-stage model heads, some efforts to enhance detection heads have been made. For good accuracy, aforementioned models need hand-designed anchors. Without the anchors which designed to be well adapted to each dataset, it was difficult to reach the same accuracy. ATSS \cite{Zhang2020ATSS} was suggested to bridge the gap between anchor-based and anchor-free models with its adaptive sampling selection algorithm. Another improved detection head add-on is RepPoints V2 \cite{chen2020reppointsv2}. RepPints V2 added head modules to compensate the inaccuracy of anchor-free detection models. However, those detection head add-ons were built with manually designed components and they have made object detection models even more complicated. Therefore, simplifying detection heads 
covering those add-on components seems still necessary.

\subsection{Vision Transformers for Object Detection}

\paragraph{Detection Transformer(DETR).} DETR \cite{DBLP:journals/corr/abs-2005-12872_DETR} is an encoder-decoder architecture that originates from the original Transformer \cite{Vaswani2017TR}. DETR has simplified the detection pipeline by removing multiple hand-designed components that encode prior knowledge, for example spatial anchors and non-maximal suppression. However, it requires additional training time(e.g. 300 epochs vs. 36 epochs of RetinaNet \cite{DBLP:journals/corr/abs-1708-02002}), and even if its performance in accuracy was good enough to be comparable to older models, e.g. Faster R-CNN, other detectors without any attention module, such as ATSS \cite{Zhang2020ATSS} and RepPoints V2 \cite{chen2020reppointsv2} showed better results than DETR in accuracy.

DETR requires very long training time to reach to the comparably moderate performance balance in accuracy and speed; however, its use of set prediction and Transformer encoder and decoder for box detection is precious legacy, which is worth visiting and reconsidering. In the Transformer module using suggested settings from their paper, the DETR encoder takes $ \frac{H_0}{32} \times \frac{W_0}{32}$  pixels from the feature maps as its query and key elements, which is a large reduction from the raw image size $H_0 \times W_0$. The DETR decoder takes the feature maps, having the same size as the encoder, as its key of attention input but uses a fixed length of object queries (e.g., 300) as its query of attention input.
\paragraph{Deformable Detection Transformer(DDETR).} Deformable DETR \cite{Zhu2020DDETR} drastically reduced the training time and enhanced the model performance of the original DETR by introducing a multi-scale deformable attention module and model redesign. As in the original DETR, the Deformable DETR encoder takes $ \frac{H_0}{32} \times \frac{W_0}{2}$ pixels in the feature maps as its key elements but not as its query elements. Instead, it takes only four, which is the default setting in the paper, sampling offsets as its query elements. Therefore, the computation complexity in the product between key and query elements is dramatically reduced from $O(H^2W^2C)$ to $O(HWC)$ when $H$,$W$ is the pixel size, $C$ is the channel size, and the other conditions follow the default settings in their paper. Deformable DETR accounts for the weaknesses of long training of the original DETR with better accuray. 

\paragraph{Dynamic Head(DyHead).} Dynamic head \cite{Dai_2021_CVPR_DyHead} combined three different self-attention mechanisms targeting feature levels for scale-awareness, spatial locations for spatial-awareness, and output channels for task-awareness by modeling the input to a detection head as a 3-dimensional tensor with dimensions level × space × channel. In comparison with Deformable DETR, DyHead better separated dimensions of inputs. However, because DyHead is still dependent on object detection frameworks, similar to ATSS \cite{Zhang2020ATSS}, which includes anchors and NMS, it could be classified as an add-on to detection models rather than an object detection model head by itself.

\subsection{Task Specific Sibling heads}

 In object detection, the output of detection models has the dual purpose of classification and localization. With single stream line of feature outputs, it has been found out that the output features are not well aligned between two tasks \cite{DBLP:journals/corr/abs-1904-06493, DBLP:journals/corr/abs-2003-07540} . To spatially disentangle the gradient flows of classification and localization, Double-Head R-CNN  \cite{DBLP:journals/corr/abs-1904-06493} and TSD  \cite{DBLP:journals/corr/abs-2003-07540} developed independent module heads for each task.
 
\begin{center}
\input{figure2}
\end{center}
\section{Method}
\subsection{Transformer Backbone}
 With the recent popularity of visual Transformers, there exist several options \cite{DBLP:journals/corr/abs-2010-11929_ViT, DBLP:journals/corr/abs-2012-12877_DeiT, Liu2021Swin} when considering the Transformer backbone for our experiment. 
 We followed the reports of \cite{Liu2021Swin} and \cite{Dai_2021_CVPR_DyHead}, which fused detection heads with the Swin Transformer and are the best combinations for good performances. 
 
 \subsection{Transformer Detection Head}
In building Transformer detection models without any specific component depending on prior knowledge about tasks, options are limited for the development is in its early stage. DETR and Deformable DETR are in a handful of options in the given constraint. Because Deformable DETR is more reliable for training and more effective for performance, this was a prior choice for our experiment.
\subsection{Task Specific Split Transformer Head}
 \paragraph{Transformer embedding} The encoder-decoder structure from the original Transformer \cite{Vaswani2017TR} was developed for the language translation task. Therefore, the output module in the Transformer has the single purpose of word embedding. However, in the case of object detection, the output module in DETR has the dual purpose of classification and localization. In DETR and Deformable DETR as well, each head module for classification and regression (localization) contains a single layer of linear projection (for classification) or a few layers of logistic regression (for localization) with the same feature input from Transformer output. As following the findings of Double-Head R-CNN  \cite{DBLP:journals/corr/abs-1904-06493} and TSD  \cite{DBLP:journals/corr/abs-2003-07540}, based on the hypothesis that feature embedding for classification and regression are not well aligned, we developed independent Transformer heads for each task.
 
\paragraph{Encoder-Decoder roles} From functional aspect, the decoder module was designed to transform spatial features into a list of detected objects. On the other hand, it is not very clear what the explicit purpose of the encoder module is. However, in the context of the object detection process, we assume that it might play a similar role to a local object feature generator or region proposal network (RPN) in the Faster R-CNN. In particular, for the Deformable DETR's 2-stage variant, the encoder's task is more explicitly targeted to assume the RPN role. However, even in this case, the decoder is involved in multiple roles, including classification, box regression, and list refinement. 

\paragraph{Task Specific Split Transformer} As illustrated in Figure \ref{fig:fig2}, we adjusted the number of layers of the Deformable DETR decoders to the half of the original model and split the output and deliver them to the classification and regression decoders. Unlike the decoders of DETR and Deformable DETR, each decoder of TSST assumes a dedicated role of classification or localization.

\paragraph{TSST class decoder}
In the Deformable DETR with an iterative bounding box refinement module, each decoder layer refines the bounding boxes based on the predictions from the previous layer, but because the TSST class decoder is not dedicated to the box prediction task, it does not need the iterative bounding box refinement add-ons, which are for box predictions. The input and output of the TSST class decoder are basically same with DETR and Deformable DETR's decoder counter parts. They have the input dimensions of $[Q; D]$, where Q is the number of object queries and $D$ is the dimension of features. It is connected to one linear layer whose output is the classification prediction with the dimensions of output $[Q; C]$ where $C$ is the class numbers, which is 80 in case of COCO. In DETR's and Deformable DETR' decoder, $Q$ is typically from 100 to 300, and $D$ is 1024. We follow the convention such that $Q$ is 300 and $D$ is 1024 in TSST class decoder. The classification prediction is paired with TSST regression decoder's output (described next) to be used for object detection set prediction loss. Please read \cite{Zhu2020DDETR} to find details about set prediction loss.

\paragraph{TSST regression decoder}
Besides the TSST class decoder, the TSST regression decoder assumes the box prediction role; thus, it does need the iterative bounding box refinement add-ons. In addition, we did not use the iterative bounding box refinement add-ons of the Deformable DETR decoder because it supports decoding roles for both classification and localization. In parallel with the TSST class decoder, the design of the TSST regression decoder's other parts are identical to the TSST class decoder except that the output of the TSST regression decoder is connected to three multi-layer perceptrons(MLP) whose output is 4 values to predict box positions and sizes. The box prediction outputs are joined with TSST class docoder's classfication, and both of them are used for set prediction loss as mentioned before.

\section{Experiments}
\subsection{Implementation Details}

\paragraph{Dataset.}
We conducted experiments on COCO object detection \cite{DBLP:journals/corr/LinMBHPRDZ14_COCO} following the commonly used settings \cite{DBLP:journals/corr/abs-2005-12872_DETR, Zhu2020DDETR}, which report AP as bounding box AP, the integral metric over multiple thresholds. COCO contains 80 categories of approximately 160K images, which are split into the train2017, val2017, and test2017 subsets having 118K, 5K, and 41K images, respectively. In all our experiments, we used only the train2017 images for training. Evaluation was performed using the val and test-dev image sets.

\paragraph{Training.} ImageNet’s pre-trained ResNet-50 was utilized as the backbone for ablations. The hyper-parameter
setting and training strategy mainly followed that of Deformable DETR. We used the following settings: multi-scale training (resizing the input such that the shorter side is between 480 and 800 while the longer side is at most 1333), AdamW optimizer (initial learning rate of 0.0002, weight decay of 0.0001, and batch size of 8), and schedule of 50 epochs. The learning rate was decayed at the 40-th epoch by a factor of 0.1.

\subsection{Ablation Study for Hyperparameters}

\begin{table}[hbt!]
	\caption{Effect of decoder size. Each row corresponds to a model with varied number of decoder and TSST layers. }
	\centering
	
    \sisetup{detect-weight,mode=text}
    \renewrobustcmd{\bfseries}{\fontseries{b}\selectfont}
    \renewrobustcmd{\boldmath}{}
    \newrobustcmd{\B}{\bfseries}
	
	\begin{tabular}{l|l|l|lll|lll}
		\toprule
		\#decoders & \#TSST layers    & \#params     & AP & AP$_{50}$ & AP$_{75}$ & AP$_S$ & AP$_M$ & AP$_L$ \\
		\midrule
		6  & 0 & 4.1M & 46.2 & 65.2 & 50.0 & 28.8 & 49.2 & 61.7 \\
		3  & 3 & 4.4M & 48.1 & \B 66.7 & 52.1 & \B 30.9 & \B 51.4 & 62.0 \\
		6  & 3 & 4.8M & \B 48.3 & 66.6 & \B 52.3 & 30.3 & 51.3 & \B 62.8       \\
		3  & 6 & 5.1M & 45.3 & 64.2 & 49.5 & 28.1 & 48.4 & 57.9        \\
		6  & 6 & 5.4M & 44.1 & 63.5 & 48.3 & 28.3 & 46.7 & 57.6  \\
		
		\bottomrule
	\end{tabular}
	\label{tab:table1}
\end{table} 

\paragraph{Optimal number of layers.}
Table \ref{tab:table1} lists the ablations for the TSST design choices for the number of TSST layers. The TSST module can replace a section of the DETR decoders or can be added to the end of the Deformable DETR. We found that the number of TSST layers needs not be as large as that of the DETR decoder. More specifically, increasing the number of TSST layers beyond three returned diminishing results. Replacing half of the DETR decoding layers with TSST layers showed promising results in terms of model size effectiveness in average precision. 

\begin{threeparttable}
	\caption{Comparison of TSST with state-of-the-art methods on COCO 2017 validation set. }
	\centering
	
    \sisetup{detect-weight,mode=text}
    \renewrobustcmd{\bfseries}{\fontseries{b}\selectfont}
    \renewrobustcmd{\boldmath}{}
    \newrobustcmd{\B}{\bfseries}
    
	\begin{tabular}{l|l|l|l|lll|lll}
		\toprule
	    Method     & Backbone & GFlops/FPS & \#params & AP & AP$_{50}$ & AP$_{75}$ & AP$_S$ & AP$_M$ & AP$_L$ \\
		\midrule
		DDETR     & ResNet-50 & 126 / 10.9\tnote{\textdagger}  &    41M    & 46.2 & 65.2 & 50.0 & 28.8 & 49.2 & 61.7 \\
		
		DDETR\_E7\_D8     & ResNet-50 & 138 / 9.6\tnote{\textdagger}  &    44M    & 46.5 & 65.8 & 50.1 & 29.2 & 49.6 & 61.5 \\
		
		DDETR\_D9     & ResNet-50 & 129 / 9.6 \tnote{\textdagger}  &    44M    & 45.2 & 64.2 & 48.9 & 27.7 & 48.2 & 61.1 \\
		    
		\midrule
        TSST(ours)  & ResNet-50     & 129 / 10.1\tnote{\textdagger}  & 44M & \B 48.1 & \B 66.7 & \B 52.1 & \B 30.9 & \B 51.4 & \B 62.0 \\
		\bottomrule
	\end{tabular}
	\begin{tablenotes}[para]
	    \footnotesize
        \item[\textdagger] For fair comparison, we tested all theoretical GFlops with 800 pixels and inference speed(FPS) results with mmdetection \cite{mmdetection} packages with V100 single GPU.
	\end{tablenotes}
	\label{tab:table2}
\end{threeparttable}

\paragraph{Parameter efficiency.}
Table \ref{tab:table2} lists the ablations for the TSST parameter efficiency compared to the original Deformable DETR. Instead of adding TSST modules to Deformable DETR, alternative design choice is to naively increase the layers of Deformable DETR. At the table, DDETR\_E7\_D8 stacks one more layer to the original Deformable DETR's six layers of encoder, and stacks two more layers upon the original six layers of decoder. Likewise, DDETR\_D9 stacks three more layers upon the original six layers of DDETR decode, but the number of layers of encoder was preserved. However, stacking extra layers upon the baseline model does not show effectiveness in either case.  On the other hand, adding the same number of parameters to TSST module reveals enhanced results in accuracy. As shown in Table \ref{tab:table1} and Table \ref{tab:table2}, increasing the number of layers in detection heads more than necessary is not a good strategy in designing optimal detection models. 

\subsection{Comparison to the State-of-the-art Methods}
\begin{threeparttable}
	\caption{Comparison of TSST with state-of-the-art methods on COCO 2017 validation set. }
	\centering
	
    \sisetup{detect-weight,mode=text}
    \renewrobustcmd{\bfseries}{\fontseries{b}\selectfont}
    \renewrobustcmd{\boldmath}{}
    \newrobustcmd{\B}{\bfseries}
    
	\begin{tabular}{l|l|l|l|lll|lll}
		\toprule
	    Method     & Backbone & GFlops/FPS & \#params & AP & AP$_{50}$ & AP$_{75}$ & AP$_S$ & AP$_M$ & AP$_L$ \\
		\midrule
		DETR & ResNet-50 &  57 / 17.6 \tnote{\textdagger} & 41M & 42.0 & 62.4 & 44.2 & 20.5 & 45.8 & 61.1 \\
		Deformable DETR     & ResNet-50 & 126 / 10.9\tnote{\textdagger}  &    41M    & 46.2 & 65.2 & 50.0 & 28.8 & 49.2 & 61.7 \\
DyHead     & ResNet-50  & -\tnote{\textdaggerdbl} &  - \tnote{\textdaggerdbl}& 42.6 & 60.1 & 46.4 & 26.1 & 46.8 & 56.0 \\
		
		\midrule
        TSST(ours)  & ResNet-50     & 129 / 10.1\tnote{\textdagger}  & 44M & \B 48.1 & \B 66.7 & \B 52.1 & \B 30.9 & \B 51.4 & \B 62.0 \\
		\bottomrule
	\end{tabular}
	\begin{tablenotes}[para]
	    \footnotesize
        \item[\textdagger] For fair comparison, we tested all theoretical GFlops and inference speed(FPS) results with mmdetection \cite{mmdetection} packages with V100 single GPU.
        \item[\textdaggerdbl] They were not provided in \cite{Dai_2021_CVPR_DyHead} and the model is not included in mmdetection packages.   
	\end{tablenotes}
	\label{tab:table3}
\end{threeparttable}

We compared our proposed TSST to Deformable DETR and DyHead \cite{Dai_2021_CVPR_DyHead}, which are considered state-of-the-art methods. As shown in Table \ref{tab:table2}, compared with Deformable DETR and DyHead, TSST achieves better performance over all object sizes. TSST's settings were similar to the original Deformable DETR and did not require extra adjustment to achieve better results. Compared with Deformable DETR, TSST requires slightly more computation. TSST needs only $ 7\% $ of more parameters and only $2.4\%$ of more computations. The training epochs were 50 which is the same with Deformable DETR's setting. The read training time for both Deformable DETR and TSST is about 3 or 4 days in our setting, which is not exactly measurable for comparison because it depends on the variety of situations. However, additional training time due to more parameters was almost negligible. 

\begin{table}[hbt!]
	\caption{Comparison of TSST with state-of-the-art methods on COCO 2017 test-dev set. }
	\centering
	
    \sisetup{detect-weight,mode=text}
    \renewrobustcmd{\bfseries}{\fontseries{b}\selectfont}
    \renewrobustcmd{\boldmath}{}
    \newrobustcmd{\B}{\bfseries}
	\begin{tabular}{l|l|lll|lll}
		\toprule
	    Method     & Backbone     & AP & AP$_{50}$ & AP$_{75}$ & AP$_S$ & AP$_M$ & AP$_L$ \\
		\midrule
		Deformable DETR     & ResNet-50       & 46.9 & 66.4 & 50.8 & 27.7 & 49.7 & 59.9 \\
		
		DyHead     & ResNet-50       & 43.0 & 60.7 & 46.8 & 24.7 & 46.4 & 53.9 \\
		DyHead     & ResNet-101       & 46.5 & 64.5 & 50.7 & 28.3 & 50.3 & 57.5 \\
		DyHead     & ResNext-64x4d-101       & 47.7 & 65.7 & 51.9 & \B 31.5 & \B 51.7 & \B 60.7 \\
		\midrule
        TSST(ours)  & ResNet-50 & \B 48.2 & \B 66.7 & \B 52.3 &  29.5 &  50.9 & 60.6\\
		\bottomrule
	\end{tabular}
	\label{tab:table4}
\end{table}

Table \ref{tab:table3} shows that our TSST performed better than Deformable DETR and DyHead on the COCO 2017 test-dev set. The performance of our TSST method with ResNet-50 surpassed DyHead, even when combined with more powerful backbones such as ResNet-101 and ResNext-101.

\begin{table}[hbt!]
	\caption{Comparison of TSST with state-of-the-art detection methods on COCO 2017 validation set. }
	\centering
	
    \sisetup{detect-weight,mode=text}
    \renewrobustcmd{\bfseries}{\fontseries{b}\selectfont}
    \renewrobustcmd{\boldmath}{}
    \newrobustcmd{\B}{\bfseries}
	\begin{tabular}{l|l|lll|lll}
		\toprule
		Method     & Backbone     & AP & AP$_{50}$ & AP$_{75}$ & AP$_S$ & AP$_M$ & AP$_L$ \\
		\midrule
		RepPoints v2 & Swin-T  & 50.0  & 68.5 & 54.2 & - & - & -  \\
		ATSS     & Swin-T & 47.2 & 66.5 & 51.3 & - & - &       \\
		Cascase Mask R-CNN     & Swin-T       & 50.4 &69.2 & 54.7 & 33.8 & 54.1  & 65.2 \\
		DyHead     & Swin-T       & 49.7 & 68.0 & 54.3 & 33.3 & 54.2 & 64.2 \\
		Deformable DETR(ours)    & Swin-T       & 49.7 & 69.0 & 54.0 & 32.5 & 52.9 & 64.7\\
		\midrule
		TSST(ours)     & Swin-T       & 51.3 & 70.3 & 55.8 &  \B 34.2 & \B 54.6 & \B 66.4 \\
		TSST(ours, test-dev)     & Swin-T       & \B 51.5 & \B 70.5 & \B 56.0 &  32.1 & 54.2 & 65.4 \\
		\bottomrule
	\end{tabular}
	\label{tab:table5}
\end{table}

Finally, we compared TSST with other models supported by the visual Transformer backbone. For the following reasons, we utilized Swin-Tiny for comparison. First, previous studies \cite{Liu2021Swin, Dai_2021_CVPR_DyHead} proved that Swin-Tiny is cost effective when compared to ResNet-50, and there are cases where extensive experiments had been performed with various detection heads, such as RepPoints V2 \cite{chen2020reppointsv2}, ATSS \cite{Zhang2020ATSS}, R-CNN, DyHead  \cite{Dai_2021_CVPR_DyHead}, etc. Table \ref{tab:table4} shows that our TSST performed better, by a large margin, than all detection heads when fused with the vision Transformer backbone. We attribute this good performance to our TSST primarily depending on Transformer modules, while the other detection heads do not.

\section{Conclusion}
In this study, we proposed the TSST method, which is an end-to-end object detector add-on with a minimal cost. Although the TSST method reflects one of the simplest ideas of introducing a task specific attention module, it also helps to fill the gap to make an anchor-free and NMS-free Transformer-based object detection method. We demonstrated that this method
achieves remarkable results compared with highly optimized competitors on the challenging COCO dataset.

However, there are still arduous challenges. Notably, we used heterogeneous (attention) methods for both the backbone and detection head. In theory, it may be sufficient to use a single homogeneous attention method for the backbone and detection head. In addition, for more advanced competitors, an improvement is necessary, in particular, regarding the combination with larger and more sophisticated backbones. We expect future work to successfully address these issues.

\paragraph{Acknowledgements}
We thank Geonmo Gu, Byungsoo Ko, and Jongtack Kim from NAVER Vision team  for discussions and advices without which this work would not be possible.

\bibliographystyle{splncs04}
\bibliography{references}

%% file: figure1.tex
\begin{figure}
	\centering
 \tikzstyle{every node}=[font=\small]
 \tikzstyle{cover} = [rectangle, minimum width=9cm, minimum height=3.5cm, draw=black,dashed, fill opacity=0, text opacity=1]
  \tikzstyle{old_od} = [rectangle, minimum width=2cm, text width=2cm, minimum height=1cm, text centered, draw=black, fill=blue!20]
 \tikzstyle{t_backbone} = [rectangle, minimum width=2cm, text width=2cm, minimum height=2cm, text centered, draw=black, fill=blue!20]
 \tikzstyle{t_od_head} = [rectangle, minimum width=2cm, text width=2cm, minimum height=2cm, text centered, draw=black, fill=blue!20]
 \tikzstyle{ranker} = [rectangle, minimum width=2cm, text width=2cm, minimum height=2cm, text centered, draw=black, fill=blue!20]
 \tikzstyle{detected} = [rectangle, minimum width=1cm, minimum height=1cm, draw=red, fill=red!5, fill opacity=0.2, text opacity=1]
 \tikzstyle{arrow} = [thick,->,>=stealth]
 \tikzstyle{dashed arrow} = [dashed, thick,->,>=stealth]

\begin{tikzpicture}[node distance=5cm]
\node[inner sep=0pt] (cat_o) at (-2,0)
    {\includegraphics[width=.15\textwidth]{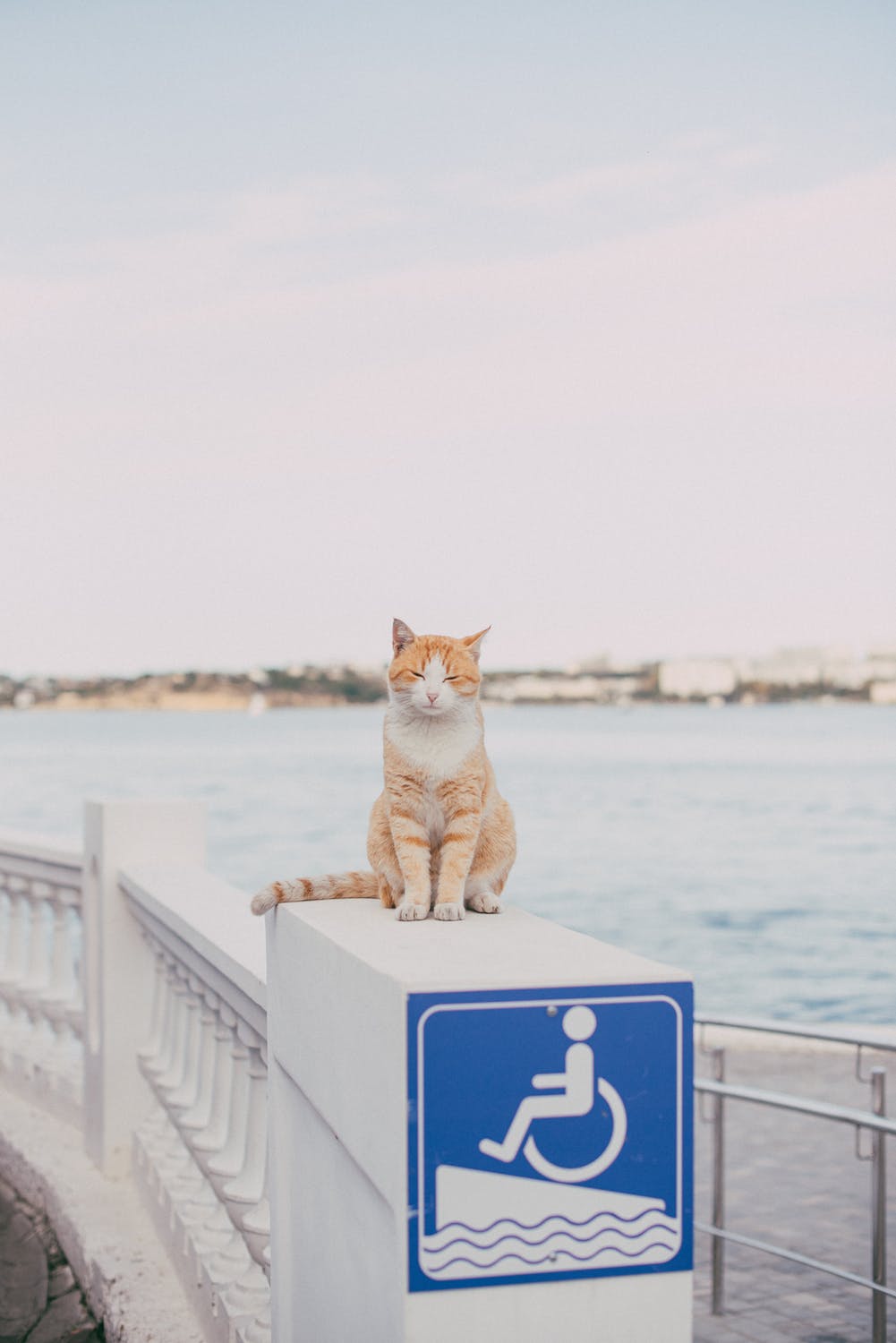}};
\node (n_cover_o) [cover] at (4, 0) {\parbox[b][3.5cm]{7cm}{\centering conventional object detector}};
\node (T_backbone_o) [t_backbone, right of = cat_o, node distance=3cm] {Backbone};
\node (neck) [old_od, right of = T_backbone_o, node distance=3cm, yshift=1.0cm] {Detection neck};
\node (anchor) [old_od, right of = neck, node distance=3cm] {anchor};
\node (head) [old_od, below of = neck, node distance=1.8cm] {Detection head};
\node (nms) [old_od, right of = head, node distance=3cm] {NMS};
\node[inner sep=0pt] (detected_cat_o) [right of = nms, node distance=3cm, yshift=0.8cm]
     {\includegraphics[width=.15\textwidth]{cat.jpg}};
\node (dcat_o) [detected, below of = detected_cat_o, node distance=0.3cm] {\parbox[b][1cm]{1cm}{Cat}};
\draw [arrow] (cat_o) -- (T_backbone_o);
\draw [arrow] (T_backbone_o) -- (neck);
\draw [arrow] (neck) -- (anchor);
\draw [arrow] (anchor) -- (head);
\draw [arrow] (head) -- (nms);
\draw [arrow] (nms) -- (detected_cat_o);

\node[inner sep=0pt] (cat_1) at (-2,-5)
    {\includegraphics[width=.15\textwidth]{cat.jpg}};
\node (n_cover_1) [cover] at (4, -5) {\parbox[b][3.5cm]{7cm}{\centering DETR / Deformable DETR}};
\node (T_backbone_1) [t_backbone, right of = cat_1, node distance=4cm] {Backbone CNN};
\node (T_od_head_1) [t_od_head, right of = T_backbone_1, node distance=4cm] {Detection Transformer};
\node[inner sep=0pt] (detected_cat_1) [right of = T_od_head_1, node distance=4cm]
     {\includegraphics[width=.15\textwidth]{cat.jpg}};
\node (dcat_1) [detected, below of = detected_cat_1, node distance=0.3cm] {\parbox[b][1cm]{1cm}{Cat}};
\draw [arrow] (cat_1) -- (T_backbone_1);
\draw [arrow] (T_backbone_1) -- (T_od_head_1);
\draw [arrow] (T_od_head_1) -- (detected_cat_1);

\node[inner sep=0pt] (cat_2) at (-2,-10)
    {\includegraphics[width=.15\textwidth]{cat.jpg}};
\node (n_cover_2) [cover] at (4, -10) {\parbox[b][3.5cm]{7cm}{\centering simplified object detection transformer}};
\node (T_backbone_2) [t_backbone, right of = cat_2, node distance=3cm] {Backbone Transformer};
\node (T_od_head_2) [t_od_head, right of = T_backbone_2, node distance=3cm] {Detection Transformer};
\node (T_ranker_2) [ranker, right of = T_od_head_2, node distance=3cm] {Task Specific Transformer};
\node[inner sep=0pt] (detected_cat_2) [right of = T_ranker_2, node distance=3cm]
     {\includegraphics[width=.15\textwidth]{cat.jpg}};
\node (dcat_2) [detected, below of = detected_cat_2, node distance=0.3cm] {\parbox[b][1cm]{1cm}{Cat}};
\draw [arrow] (cat_2) -- (T_backbone_2);
\draw [arrow] (T_backbone_2) -- (T_od_head_2);
\draw [arrow] (T_od_head_2) -- (T_ranker_2);
\draw [arrow] (T_ranker_2) -- (detected_cat_2);
\end{tikzpicture}
	\caption{: Illustration of the conventional object detection vs. simplified object detection transformer}
	\label{fig:fig1}
\end{figure}

%% file: figure2.tex
\begin{figure}
	\centering
\usetikzlibrary{shapes.geometric, arrows}
\tikzstyle{every node}=[font=\small]
\tikzstyle{cover} = [rectangle, minimum width=4cm, minimum height=3.5cm, draw=gray,dashed, fill opacity=0, text opacity=1]
\tikzstyle{encoder} = [rectangle, minimum width=2cm, text width=2cm, minimum height=2cm, text centered, draw=black, fill=orange!30]
\tikzstyle{decoder} = [rectangle, minimum width=2cm, text width=2cm, minimum height=2cm, text centered, draw=black, fill=green!30]
\tikzstyle{c_decoder} = [rectangle, minimum width=2cm, text width=2cm, minimum height=2cm, text centered, draw=black, fill=yellow!30]
\tikzstyle{r_decoder} = [rectangle, minimum width=2cm, text width=2cm, minimum height=2cm, text centered, draw=black, fill=purple!30]
\tikzstyle{tt_backbone} = [trapezium, trapezium left angle=80, trapezium right angle=80, minimum width=2cm, minimum height=4cm, text centered, draw=black, fill=pink!30]
\tikzstyle{queries} = [rectangle, minimum width=2.5cm, text width=3cm, minimum height=1cm, text centered, draw=black, fill=blue!30]
\tikzstyle{textbox} = [rectangle, minimum width=3cm, minimum height=1cm, text centered] 
\tikzstyle{arrow} = [thick,->,>=stealth]
\tikzstyle{dashed arrow} = [dashed, thick,->,>=stealth]

\begin{tikzpicture}[node distance=4cm]
trapezium stretches=true %
\node (encoder_cover) [cover, xshift=0.6cm, yshift=0.6cm] {};
\node (encoder1) [encoder] {};
\node (encoder2) [encoder, xshift=0.2cm, yshift=0.2cm] {};
\node (encoder3) [encoder, xshift=0.4cm, yshift=0.4cm] {};
\node (encoder4) [encoder, xshift=0.6cm, yshift=0.6cm] {};
\node (encoder5) [encoder, xshift=0.8cm, yshift=0.8cm] {};
\node (encoder6) [encoder, xshift=1.0cm, yshift=1.0cm] {Deformable DETR encoder};
\node (tt_backbone) [tt_backbone, left of = encoder1, node distance=3cm, anchor=base, inner xsep=-0.2cm, inner ysep=1.5cm] {backbone};
\node (decoder_cover) [cover, right of = encoder6, xshift=0.3cm, yshift=0.3cm] {};
\node (decoder1) [decoder, right of = encoder6, yshift=0.0cm] {};
\node (decoder2) [decoder, right of = encoder6, xshift=0.2cm, yshift=0.2cm] {};
\node (decoder3) [decoder, right of = encoder6, xshift=0.4cm, yshift=0.4cm] {Deformable DETR decoder};
\node (object queries) [queries, below of = decoder1, node distance=4cm] {object queries};
\node (c_decoder_cover) [cover, right of = decoder3, xshift=0.3cm, yshift=2.3cm] {};
\node (class decoder1) [c_decoder, right of = decoder3, yshift=2cm] {};
\node (class decoder2) [c_decoder, right of = decoder3, xshift=0.2cm, yshift=2.2cm] {};
\node (class decoder3) [c_decoder, right of = decoder3, xshift=0.4cm, yshift=2.4cm] {TSST class decoder};
\node (r_decoder_cover) [cover, below of = class decoder1, xshift=0.3cm, yshift=0.3cm, node distance=4cm] {};
\node (regression decoder1) [r_decoder, below of = class decoder1, node distance=4cm] {};
\node (regression decoder2) [r_decoder, below of = class decoder1, node distance=4cm, xshift=0.2cm, yshift=0.2cm] {};
\node (regression decoder3) [r_decoder, below of = class decoder1, node distance=4cm, xshift=0.4cm, yshift=0.4cm] {TSST regression decoder};
\node (c_pred) [textbox, above of = class decoder3, node distance=1.5cm] {classification predictions};
\node (r_pred) [textbox, above of = regression decoder3, node distance=1.5cm] {bounding box predictions};
\node (image) [textbox, below of = tt_backbone, node distance=3cm] {image};
\node (image feature maps) [textbox, above of = tt_backbone, node distance=3.0cm, xshift=0cm] {image feature maps};
\node (arrow_exp) [textbox, below of = object queries, node distance=3cm, xshift=2cm] {2-stage ROI};

\draw [arrow] (image) -- (tt_backbone);
\draw [-, thick] (tt_backbone.north) -|++(0,0.5) -|++(1.0,0) -|([xshift=5]tt_backbone.east);
\draw [arrow] ([xshift=5]tt_backbone.east) to[right] node[auto] {} ++ (0.5,0);
\draw [arrow] (encoder6) -- (decoder1);
\draw [arrow] (object queries) -- (decoder1);
\draw [arrow] (decoder3) -- (class decoder1);
\draw [arrow] (decoder3) -- (regression decoder1);
\draw [dashed arrow] (encoder6) -- (object queries);
\draw [dashed arrow] (arrow_exp) to[left] node[auto] {} ++ (-3,0);
\end{tikzpicture}
	\caption{: Illustration of the proposed TSST object detector.
}
	\label{fig:fig2}
\end{figure}
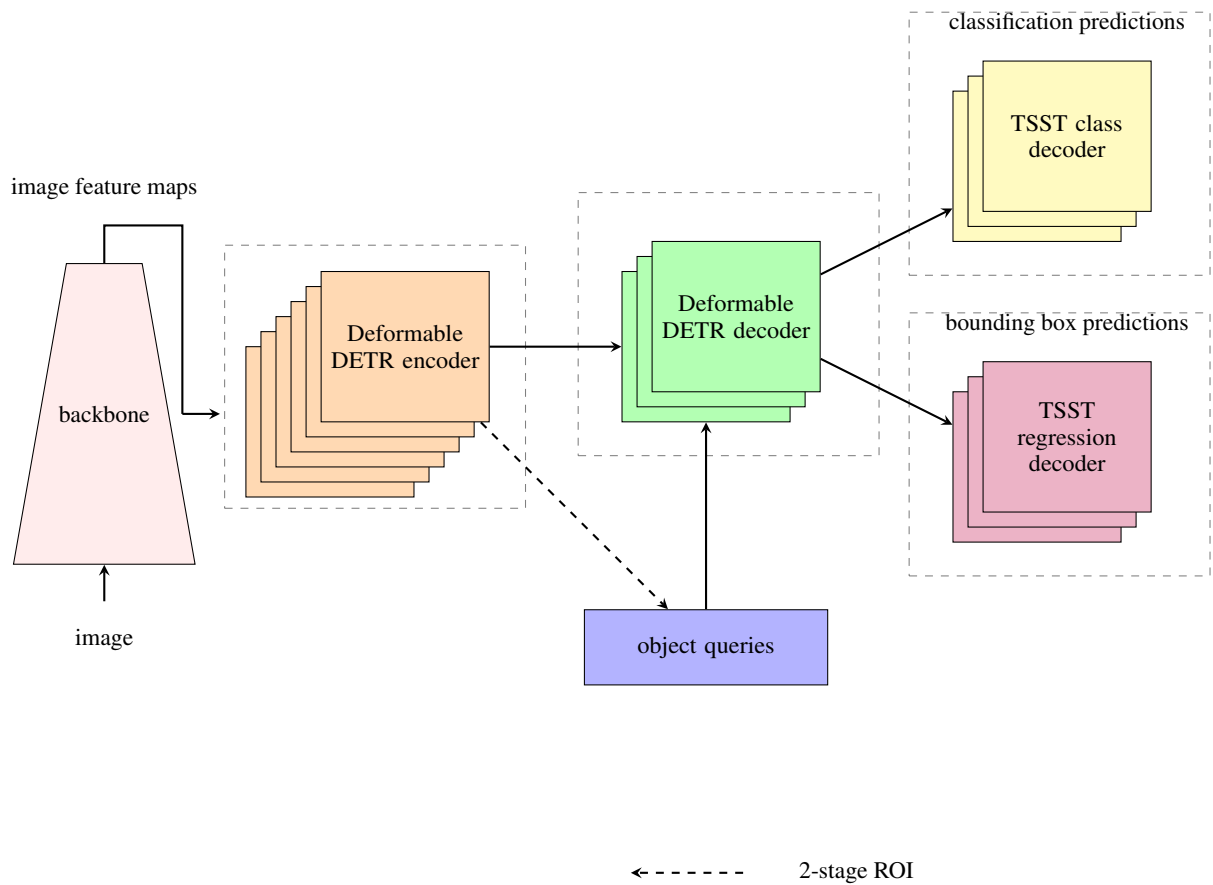